\documentclass[conference,a4paper]{APSIPA2021}
\usepackage{amsmath,amssymb,bm}
\usepackage{graphicx}
\usepackage[caption=false,subrefformat=parens,font=footnotesize]{subfig}
\usepackage[ruled]{algorithm2e}
\usepackage{hyperref,url}
\usepackage{booktabs}
\usepackage[style=ieee,maxbibnames=6]{biblatex}

\hypersetup{
    colorlinks,
    urlcolor=blue
}
\newcommand{\aref}[1]{\hyperref[#1]{Appendix~\ref*{#1}}} % to reference appendix sections as Appendix rather than Section as autoref doesn't do it
\newcommand{\ra}[1]{\renewcommand{\arraystretch}{#1}} % stretch the spacing between rows  in tables
\newcommand{\x}{\mathbf{x}} % bold x for vectors
\newcommand{\p}{\mathbf{p}} % bold p for vectors
\newcommand{\z}{\mathbf{z}} % bold z for vectors
\newcommand{\bu}{\mathbf{u}} % bold u for vectors
\newcommand{\e}{\mathbf{e}} % bold e for vectors
\newcommand{\bc}{\mathbf{c}} % bold c for vectors

\begin{document} 

\title{Landmark Management in the Application of Radar SLAM}

\author{\authorblockN{%
Shuai Sun\authorrefmark{1} and Beth Jelfs\authorrefmark{2} and Kamran Ghorbani\authorrefmark{3} and Glenn Matthews\authorrefmark{3} and Christopher Gilliam\authorrefmark{2}
}
\authorblockA{%
\authorrefmark{1}
Navigation College,  Dalian Maritime University, Liaoning, China \\
E-mail: shuai.sun@dlmu.edu.cn}
\authorblockA{%
\authorrefmark{2}
Dept. Electronic, Electrical \& Systems Engineering, University of Birmingham, Birmingham, UK\\
E-mail: \{b.jelfs, c.gilliam.1\}@bham.ac.uk}
\authorblockA{%
\authorrefmark{3}
School of Engineering, RMIT University, Melbourne, Australia\\
E-mail: \{kamran.ghorbani, glenn.matthews\}@rmit.edu.au}
}

\maketitle
\thispagestyle{empty}

\begin{abstract}
This paper focuses on efficient landmark management in radar based simultaneous localization and mapping (SLAM). Landmark management is necessary in order to maintain a consistent map of the estimated landmarks relative to the estimate of the platform's pose. This task is particularly important when faced with multiple detections from the same landmark and/or dynamic environments where the location of a landmark can change. A further challenge with radar data is the presence of false detections. Accordingly, we propose a simple yet efficient rule based solution for radar SLAM landmark management. Assuming a low-dynamic environment, there are several steps in our solution: new landmarks need to be detected and included, false landmarks need to be identified and removed, and the consistency of the landmarks registered in the map needs to be maintained. To illustrate our solution, we run an extended Kalman filter SLAM algorithm in an environment containing both stationary and temporally stationary landmarks. Our simulation results demonstrate that the proposed solution is capable of reliably managing landmarks even when faced with false detections and multiple detections from the same landmark.
\end{abstract}

%===============================================================================
\section{Introduction}
\label{sec:Intro}
%===============================================================================
Simultaneous localization and mapping (SLAM) aims to estimate the pose (location and heading) of a mobile platform, whilst simultaneously sensing the surrounding environment. SLAM in vehicular based applications is an active field of research~\cite{SLAM_past_future_2016,Li2017FusionSLAMVehicles}. The design of a SLAM algorithm is closely related to the modeling of the environment. Depending on the specific environment, as well as the types of onboard sensor employed, there are two main approaches to modeling the surrounding environment. One is a landmark based formulation~\cite{Dissanayake2002MapManagement_Robots}, where the sensors are used to detect surrounding objects which are then abstracted as landmarks, with the goal to simultaneously estimate the location of these landmarks. The second approach is an occupancy grid based formulation~\cite{Grisetti2007Grid_mapping_SLAM_TransRobotics}, where the surrounding environment is segmented into a number of disjoint occupancy grids, and the task of environment mapping is conducted by estimating whether each occupancy grid is occupied or not. 
 
In the past few decades, researchers have mainly focused on developing algorithms and conducting practical experiments on Lidar based SLAM or visual based SLAM~\cite{SLAM_survey_2016}. However, these approaches can have limitations in harsh weather conditions, hence more recently there has been increased interest in SLAM using radar sensors which appear to be better suited to these conditions, especially with rain or fog~\cite{hong2021radar_SLAM}. Applying high resolution radar sensors for SLAM can have its own challenges, especially regarding measurement modeling. Firstly, unlike many existing point landmark based SLAM approaches, landmarks of a certain size contain multiple scattering/reflecting points and thus return multiple radar detections~\cite{lee2020EKF_SLAM_TAES}. Secondly, the aspect angle between the mobile platform and the landmarks varies during the motion of the mobile platform, leading to varying number of radar detections returned for the same landmark. Finally, radar detections may occur due to clutter rather than true landmarks. Thus, an effective landmark management scheme requires the ability to reject spurious measurements at landmark initialization, and remove false landmarks from the system. 

In addition to the complexity in the radar sensor data, SLAM in a dynamic environment proves to be more challenging than in a static environment~\cite{Vu2011gridSLAM}. Most existing works on SLAM assume either a stationary environment or the capability to discriminate between static and dynamic features~\cite{Lee2013PhD_slam}. A common strategy is to eliminate the moving objects and only use static objects to update the pose of the mobile platform~\cite{2021Landmark_elimination}. However, this approach does not account for low-dynamic environments, that is, one in which landmarks can be stationary or temporally stationary~\cite{Wang2003SLAMwithDTMO}. For example, in a car park, most vehicles are stationary however some will enter and exit the parking area. In this setting, landmarks that have been confirmed and registered in the system should be removed if their positions are altered. 

In this paper, we propose a rule based landmark management scheme for SLAM in low-dynamic environments using radar sensors. We aim to produce a consistent and efficient landmark management scheme that can be applied to existing SLAM solutions. We provide a detailed illustration regarding the key steps required to confirm/initialize new landmarks, associate landmarks with multiple radar detections, remove landmarks and merge landmarks. To illustrate the proposed landmark management scheme we choose the extended Kalman filter (EKF) SLAM algorithm~\cite{Guivant2001optimization_EKF_SLAM}. We evaluate our management scheme by simulating a mobile platform with a frequency-modulated continuous-wave (FMCW) radar sensor moving around a typical car park environment. The scene contains a number of parked vehicles as landmarks and the locations for some of the landmarks in the scene may change. The simulation results show that our management scheme is robust to false radar detections, multiple detections from a single landmark and landmarks changing position.

This paper is organized as follows: in \autoref{sec:formulation} we formulate the FMCW Radar SLAM problem. \autoref{sec:solution} provides our detailed rule-based landmark management scheme. Simulation results are presented in \autoref{sec:experiment} and \autoref{sec:conclusion} concludes the work.

%===============================================================================
\section{Problem Formulation}
\label{sec:formulation}
%===============================================================================
Following the formulation of the EKF-SLAM~\cite{Guivant2001optimization_EKF_SLAM}, we define an augmented state vector at time $k$, denoted by $\x^a_k$, as
\begin{equation}
	\x_k^a = \Big[\underbrace{x_k,\ y_k,\ \theta_k}_{\x_k^m},\ \underbrace{\p_k^1,\ldots, \p_k^{N_k}}_{\x_k^{\ell}}\Big]^T,
\end{equation}
where $\x_k^m=\left[x_k,\ y_k,\ \theta_k\right]^T$ represents the pose of the mobile platform, namely the 2D location and heading of the mobile platform, and $\x_k^{\ell} = \left[\p^1_k, \ldots, \p_k^{N_k}\right]^T$ is the vector of landmarks registered in the system, with each $\p_k^n = \left[p_k^n(x),\ p_k^n(y)\right]$ representing the 2D global location of landmark $n$. Note that the total number of landmarks $N_k$ will vary with time as landmarks are added and removed.

\begin{figure}[tb]
	\centering
	\includegraphics[width=\linewidth]{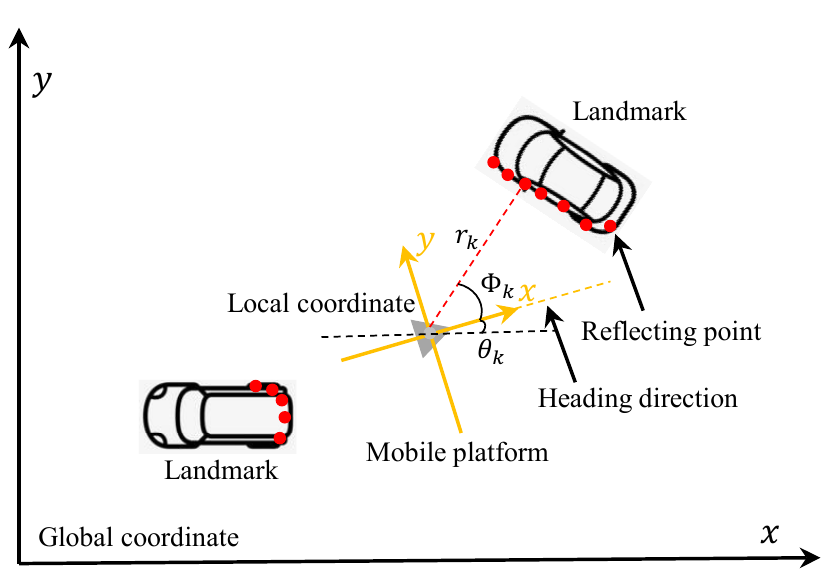}
	\caption{Illustration of a typical considered SLAM scene.}
	\label{fig:scheme}
\end{figure}

In order to identify landmarks, we assume the platform has a FMCW radar to sense its surrounding environment. This sensor measures the range and azimuth direction of a landmark relative to the platform's pose. Realistically, however, each landmark may produce multiple radar detections the number of which varies with the aspect angle between the mobile platform and the landmark~\cite{Car_tracking_2014}. An example of this scenario is shown in \autoref{fig:scheme} where the landmarks are exemplified by vehicles. The sensor measurements obtained in this scenario can be represented by using a reflecting point data model~\cite{Granstrom2016ExtendedTargetTracking_Arxiv}. Thus, a radar measurement $\z_k = [r_k,\ \phi_k]^T$ obtained at time $k$ contains the relative range, $r_k$, and azimuth, $\phi_k$, that correspond to a certain reflecting point $\left(x_n^i,\ y_n^i\right)$ from landmark $n$ such that 
\begin{equation}
    \z_k = h_n\left(\x_k^a\right) + \bm{\xi}_k, 
\end{equation}  
where
\begin{align}
    h_n(\x_k^a) = \begin{bmatrix}
	    \sqrt{\left(x_k - x_n^i\right)^2 + \left(y_k - y_n^i\right)^2 }\ \\ 
	    \arctan\left(\frac{y_k - y_n^i}{x_k - x_n^i}\right) - \theta_k
	\end{bmatrix},
\end{align}
and $\bm{\xi}_k \sim \mathcal{N}(0,R_k)$ is additive zero mean Gaussian noise with covariance matrix, $R_k$. We assume the sensor has a 360 degree field of view but is limited to a maximum detection range~\cite{VivetFMCW_SLAM_2013}.

%===============================================================================
\section{Landmark Management}
\label{sec:solution}
%===============================================================================
In this section, we present our landmark management scheme under the recursive EKF framework. Since each landmark may produce multiple radar detections, the total number of detection points can, in practice, be very large. Hence, an efficient landmark management scheme which does not significantly increase the computational load is required. If we consider the radar point cloud at time point $k$, denoted by $Z_k$ then there are three sources for the data:
\begin{itemize}
	\item from existing objects that have already been registered as landmarks in the state vector $\x_k^a$;
	\item from objects that have not been registered in the state vector $\x_k^a$;
	\item from false detections.
\end{itemize}
Hence, when processing the radar detections, $Z_k$, the first step is data association between the registered landmarks and the received radar measurements, where each landmark may be associated with multiple radar point detections. The associated measurements can then be used to update the system state. Next any remaining detections which have not been associated with an existing landmark are used to determine if any new landmarks are to be initialized. 

In the following we present a detailed description of our landmark management scheme for a single iteration of the EKF-SLAM. The main processing consists of the following three modules: state prediction; state update; landmark removal and new landmark confirmation/inclusion. We provide a complete summary of the steps in this process in \autoref{alg:ekf_slam}.

\begin{algorithm}[tb]
	\caption{Landmark management for EKF SLAM}
	\label{alg:ekf_slam}
	\KwIn{Initial pose of the mobile platform, $\x_0^m =[ x_0,\ y_0,\ \theta_0]$, Model parameters in \autoref{table:para}, Radar detections $Z_k$, Odometer reading $\bar{\bu}_k$}
	\KwOut{System state $\x_{k|k}^a, P_{k|k}^a$}
	
	Recursion:
	\For {$k=1; k < K; k++$}{
		EKF Prediction $\rightarrow$ $\left(\x_{k|k-1}^a,\ P_{k|k-1}^a\right)$\\ 
		Compute the predicted state based on~\eqref{eq:ekf_predict1}\; 
		EKF Update $\rightarrow$ $\left(\x_{k|k}^u,\ P_{k|k}^u\right)$ \\ 
		Apply ``Sifting'' to select $Z_k^s$ from $Z_k$\;
		\For {$\p_k^n \in \x_{k|k-1}^a$} {
		    Associate $\p_k^n$ with $Z_k^s$ using~\eqref{eq:distance}\;
		    Update $\x_{k|k-1}^a, P_{k|k-1}^a$ using~\eqref{eq:ekf_update}\;
		    Apply landmark removal logic $\rightarrow \left(\x_{k|k}^r, P_{k|k}^r\right)$ \\
		    \If{$\p_k^n$ \text{is confirmed to be removed}} {
			    Remove the corresponding entries in $\x_{k|k}^u,\ P_{k|k}^u$\;
		    }
		} 
		New landmark Inclusion $\rightarrow$ $\left(\x_{k|k}^{a,N_k^{\ell}},\ P_{k|k}^{a,N_k^{\ell}}\right)$ \\
		Cluster radar detections in $Z_k^r$\;
		Apply landmark confirmation for each cluster using Rule 1 and Rule 2\;
		\If{$\bc_k^j$ \text{is confirmed}} {
		    Include $\bc_k^j$ to the system using~\eqref{eq:include_1}--\eqref{eq:include_2}\;
	    }
	    Landmark Merging $\rightarrow$ $\left(\x_{k|k}^{a},\ P_{k|k}^{a}\right)$ \\
	    Apply landmark merging for the registered landmarks\;
	}
\end{algorithm}

%-------------------------------------------------------------------------------
\subsection{EKF Prediction}
\label{ssec:EKFPred}
%-------------------------------------------------------------------------------
Before performing the update of the system using the radar detections we first perform prediction of the system state given the filtered augmented state vector from the previous time step, $\x_{k-1|k-1}^a$, and the state covariance matrix, $P_{k-1|k-1}^a$. The prediction is implemented based on a motion model of the mobile platform, the odometer data, as well as the assumption that the position of landmarks remains static. The motion of the mobile platform is modeled by a state transition equation~\cite{Fazekas2021OdometryModelVehicle}
\begin{equation}
	\x^m_k = f_m\left(\x^m_{k-1},\ \bu_k\right) + \bm{\omega}_k,
\end{equation}
where $\bu_k$ is the control input for the mobile platform, and $\bm{\omega}_k \sim \mathcal{N}(\bm{\omega_}k; 0, Q_k)$ is the zero mean Gaussian process noise with covariance matrix $Q_k$. The control input is defined as $\bu_k=\left[v_k,\ \psi_k\right]$ where $v_k$ is the velocity and $\psi_k$ is the yaw rate, such that the transition of each component can be expressed as 
\begin{align}
	\begin{split}	
		x_k & = x_{k-1} +  v_k \Delta t \cos \left(\theta_{k-1} + \frac{\Delta t \psi_k}{2}\right), \\
		y_k  & = y_{k-1} +  v_k \Delta t \sin \left(\theta_{k-1} + \frac{\Delta t \psi_k}{2}\right), \\
		\theta_k & = \theta_{k-1} +  \psi_k \Delta t,
	\end{split}
\end{align} 
where  $\Delta t$ is the time interval between time $k-1$ and time $k$.

In reality, the control inputs are obtained from wheel rotation sensors, which provide noisy measurements of the true control inputs. Therefore the sensor readings of the velocity and yaw rate are denoted by $\bar{\bu}_k = \left[\bar{v}_k,\ \bar{\psi}_k\right]$, and modelled as
\begin{equation}
    \bar{\bu}_k = \bu_k + \bm{\eta}_k,
\end{equation}
with $\bm{\eta}_k \sim \mathcal{N}(\bm{\eta}_k; 0,U_k)$ a zero mean Gaussian noise and $U_k$ the corresponding covariance matrix. Given these noisy odometer measurements, the EKF prediction from time $k-1$ to time $k$ is~\cite{Bailey2002SLAM_fundamentals}
\begin{align} 
    \label{eq:ekf_predict1}
	\begin{split}
	    \x_{k|k-1}^a &= f\left(\x_{k-1|k-1}^a,\ \bar{\bu}_k\right) = \begin{bmatrix}
	 	    f_m\left(\x_{k-1|k-1}^m,\ \bar{\bu}_k\right) \\ 
	 	    \x_{k-1|k-1}^{\ell}
	    \end{bmatrix},  \\
 	    P_{k|k-1}^a &= \nabla f_{\x_k^a} P_{k-1|k-1}^a \nabla f_{\x_k^a}^T + \nabla f_{\bu_k} U_k \nabla f_{\bu_k}^T + Q_k,
 	\end{split}
\end{align}
with detailed definitions provided in \aref{ap:ekf_predict}.

%-------------------------------------------------------------------------------
\subsection{EKF Update}
\label{ssec:EKFUpdate}
\begin{figure}[tb]
    \centering
    \includegraphics[width=\linewidth]{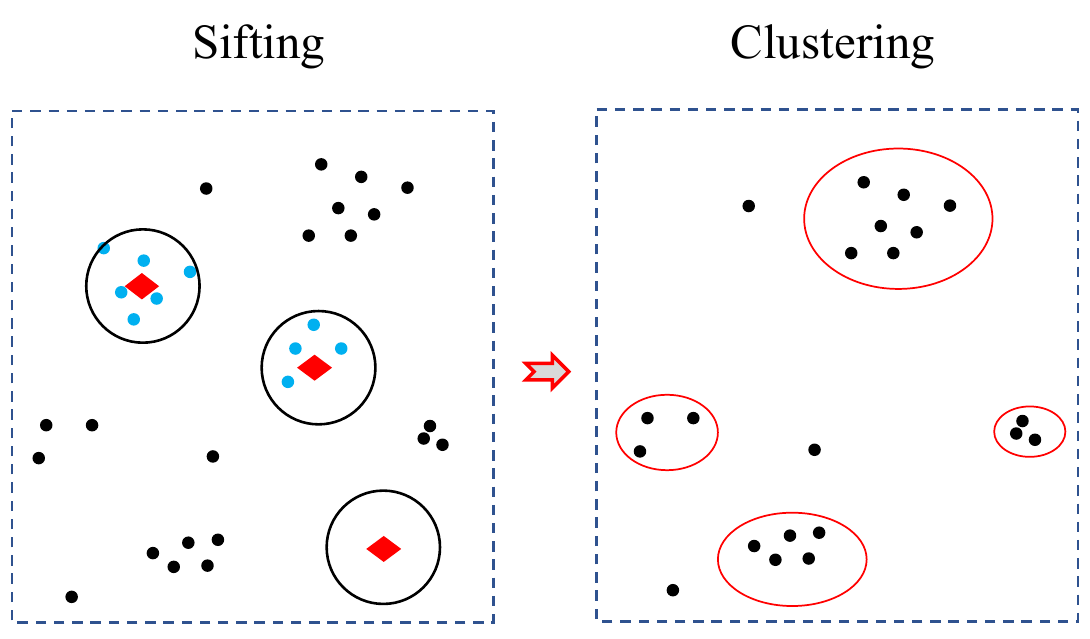}
    \caption{Illustration of sifting (on the left); each red diamond shape represents a registered landmark in the system. The remaining measurements (on the right) are clustered for the subsequent new landmark identification and inclusion. Measurements that do not belong to a cluster are treated as false detections and removed.}
    \label{fig:sifting}
\end{figure}

%-------------------------------------------------------------------------------
Before updating the predicted state vector, the association between the predicted registered landmarks and the radar detections is conducted. To alleviate the computational burden of performing association on all of the measurements, a ``sifting'' step is used as in~\cite{lee2020EKF_SLAM_TAES} to coarsely select \textit{plausible} detections as candidate measurements. The candidate measurements are identified by calculating the Euclidean distances between the measurements and the existing landmarks. Any measurements which fall within a circle of pre-defined radius, $\gamma_s$, of an existing landmark are considered as a candidate measurement, as illustrated in \autoref{fig:sifting}.
 
The candidate measurement data identified during the sifting procedure is denoted as $Z_k^s$. This measurement data is then further refined using a stricter association step. This step is based on the log likelihood distance between a given detection $\z_k$ and landmark $n$: 
\begin{equation} 
    \label{eq:distance}
    D_n = -\log \Lambda \left(\z_k | \x_{k|k-1}^m,\ \p_{k-1}^n\right),
\end{equation}
where the likelihood is defined in \aref{ap:ekf_update}. If $D_n$ is less than a threshold, $\beta$, the measurement is associated with the landmark. Once a radar measurement is associated with a registered landmark, it is used to update the system state via the standard EKF update formula:
\begin{align} 
    \label{eq:ekf_update}
    \begin{split}	
        \x_{k|k}^{a,u} & = \x_{k|k-1}^a + W_k\left(\z_k - h_n\left(\x_{k|k-1}^a\right)\right),\\
 	    P_{k|k}^{a,u} & = P_{k|k-1}^a - W_k S_k W_k^T,
    \end{split}
\end{align} 
where the corresponding definitions and implementation are provided in \aref{ap:ekf_update}.

%-------------------------------------------------------------------------------
\subsection{Landmark Removal}
\label{ssec:landmarkRemoval}
\begin{figure}[tb]
    \centering
    \includegraphics[width=\linewidth]{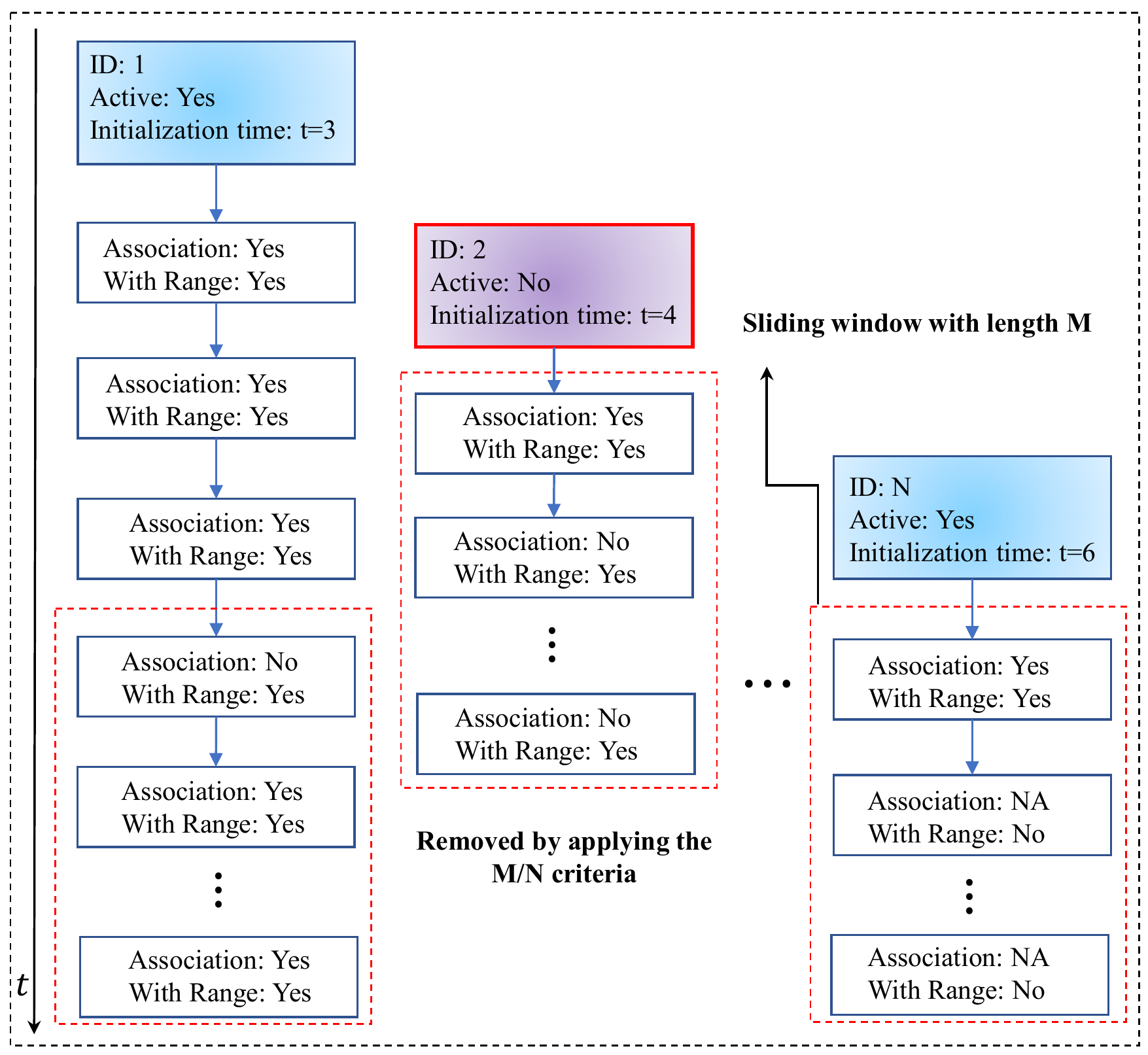}
 	\caption{Illustration of applying an M/N logic for landmark removal.}
 	\label{fig:delete}
 \end{figure}

%-------------------------------------------------------------------------------
After completing the update we next need to consider the fact that, in a slowly dynamic environment, there may be previously registered landmarks that have since changed their location. To address this problem we propose to use an $M/N$ logic based scheme to identify and remove landmarks that no longer exist from the registered list. The rationale of this logic is that landmarks that currently exist, and are within the radar detection range, should produce consistent radar detections. Using a sliding window of length $M$ we determine, based on the landmark and radar data association results from the EKF update step, if a registered landmark is associated with radar detections at least $N$ times. If a landmark is not associated with radar data at least $N$ times in the sliding window, and provided the registered landmark is still within the radar detection range, this landmark is identified for removal from the system state vector. \autoref{fig:delete} shows a detailed illustration of the landmark removal procedure using the $M/N$ logic. 
 
Note that the landmark removal procedure is conducted after the EKF update, hence, the number of landmarks immediately after the EKF update is the same as the number at the beginning of the EKF prediction. If there are landmarks to be removed, the corresponding entries in $\x_{k|k}^{a, u}$ and $P_{k|k}^{a, u}$ will be removed. This leads to a landmark reduction in the system, from $N_{k-1}$ to $N_k^r$ $\left(N_k^r \leq N_{k-1}\right)$ and correspondingly the cardinality of the system is $3+2N_k^r$. We denote the system state after landmark removal as $\x_{k|k}^{a, r}$ and $P_{k|k}^{a, r}$. 

%-------------------------------------------------------------------------------
\subsection{New Landmark Inclusion}
\label{ssec:landmarkInclusion}
%-------------------------------------------------------------------------------
After completing the removal of any landmarks that no longer exist, we next consider if there are any new landmarks to be included. There are two key issues regarding landmark inclusion: confirming it is a new landmark, and then registering this landmark in the system state. Intuitively, a new landmark must be far apart from the registered landmarks and, to avoid including any potential landmarks which may have originated from false radar detections, must produce consistent radar detections. To identify any potential new landmarks the first step is to cluster the radar detections, $Z_k^r$, which are those detections from $Z_k$ which are not included in $Z_k^s$, such that $Z_k = \Big\{Z_k^s,\ Z_k^r \Big\}$. Clustering is performed using DBSCAN~\cite{Ester96adensity-based}, with the minimum number of detections required to form a cluster determined by the threshold $N_{c_2}$ and the maximum distance between two detections for them to be considered part of the same cluster determined by the threshold $\gamma_c$, as shown in \autoref{fig:sifting}. Identified clusters are represented by their centers, corresponding to the detection with the largest return, such that for the $j^{\text{th}}$ cluster the center is $\bc_k^j=\left[r_k^j,\ \phi_k^j\right]$. The clusters are then assessed to determine whether they should be confirmed as a new landmark and registered in the system.

\begin{figure}[tb]
    \centering
    \includegraphics[width=\linewidth]{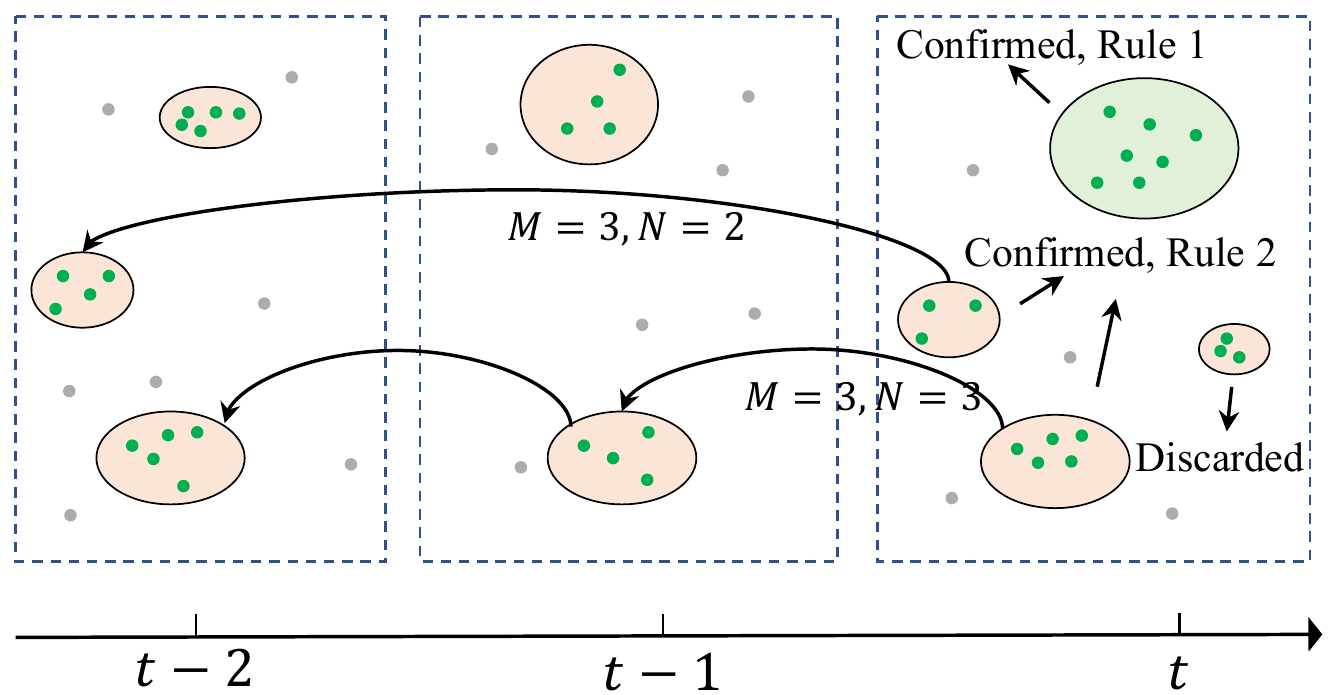}
 	\caption{Illustration of new landmark identification based on Rule 1 and Rule 2 using a 3/2 logic; note that the clustering results are from the remaining radar detections after sifting in \autoref{fig:sifting}. }
 	\label{fig:initialization}
 \end{figure}

We propose a rule based scheme for confirmation of clusters as new landmarks. First, we confirm that the cluster center is sufficiently far from any existing landmarks by calculating the log likelihood distance defined in~\eqref{eq:distance} of the cluster center with respect to the registered landmarks. If the distance exceeds a pre-defined threshold, $\alpha$, we progress to determining if it meets either of the following rules:
\begin{itemize}
    \item \textbf{Rule 1} if the number of radar detections in a cluster exceeds threshold $N_{c_1}$, it is identified as a landmark;
 	\item \textbf{Rule 2} if the number of radar detections is less than $N_{c_1}$, but it meets a multi-frame based criteria, as described below, it is identified as a landmark.
\end{itemize}
The multi-frame criteria in Rule 2 first associates current clusters with clusters from the previous time step based on the Euclidean distances between the cluster centers being less than the threshold $\gamma_a$. To be confirmed as a new landmark $M/N$ logic is used, i.e., during $M$ consecutive time steps, if there are at least $N$ times that this cluster is detected/observed, it is confirmed as a new landmark. \autoref{fig:initialization} further illustrates the rules applied for landmark confirmation.
 
Once a landmark is confirmed it needs to be initialized in the system. If there are $N_k^{\ell}$ new landmarks to be initialized, then the 2D location of the $j^{\text{th}}$ landmark $\left(1 \leq j \leq N_k^{\ell}\right)$ is %given by
% \begin{align}
%     \label{eq:include_1}
%     \begin{bmatrix}
%  		p_k^{N_k^r+ j}(x) \\ 
%  		p_k^{N_k^r + j}(y)
%  	\end{bmatrix} = \begin{bmatrix}
%  	    x_{k|k} + r_k^j \cos\left(\theta_{k|k} + \phi_k^j\right) \\ 
%  	    y_{k|k} + r_k^j \sin\left(\theta_{k|k} + \phi_k^j\right) 
%  	\end{bmatrix}.
%  \end{align}
% The transpose of~\eqref{eq:include_1} gives 
$\p_k^{N_k^r + j}$ and is concatenated to the end of the state vector to give the updated state vector 
\begin{equation}
    \label{eq:include_1}
    \x_{k|k}^{a,j} = \Big[\underbrace{\x_{k|k}^{a,r},\ \p_k^{N_K^r +1},\ldots, \p_k^{N_k+j-1}}_{\x_{k|k}^{a,j-1}},\ \p_k^{N_k^r + j}\Big]^T.
\end{equation}
The corresponding covariance matrix is then updated to give
\begin{equation}  
    \label{eq:include_2}
	P_{k|k}^{a,j} = J_1 P_{k|k}^{a, j-1} J_1^T + J_2 R_k J_2^T,
\end{equation}
and derivations of~\eqref{eq:include_1} and~\eqref{eq:include_2} are provided in \aref{ap:inclusion}~\cite{Bailey2006SLAM}. At time $k$ before landmark inclusion there are $N_k^r$ landmarks, after landmark inclusion the total number of landmarks are $N_k = N_k^r + N_k^{\ell}$ and the system state is denoted as $\x_{k|k}^{a, N_k^{\ell}}$ and $P_{k|k}^{a, N_k^{\ell}}$. 

%-------------------------------------------------------------------------------
\subsection{Landmark Merging}
\label{ssec:landmarkMerge}
%-------------------------------------------------------------------------------
Finally, the last step in our landmark management scheme is to address the possibility that two registered landmarks in the system may correspond to the same physical object\footnote{Based on our observations, this usually occurs at the clustering step, when measurements originating from the same object (typically with a large extent) are clustered into more than one cluster.}. We address this possibility by introducing a landmark merging logic step into our management scheme. The logic merges two landmarks if the Euclidean distance between the two is less than a threshold $\gamma_m$. The system state at the completion of landmark merging at time $k$ is denoted as $\x_{k|k}^{a}$ and $P_{k|k}^{a}$ which completes the single iteration of \autoref{alg:ekf_slam}.

%===============================================================================
\section{Simulation Results}
\label{sec:experiment}
\begin{figure*}[htp] 
\centering
    \subfloat[$k=2$]{\includegraphics[clip, trim=7mm 1mm 14mm 6mm,width=0.4\linewidth]{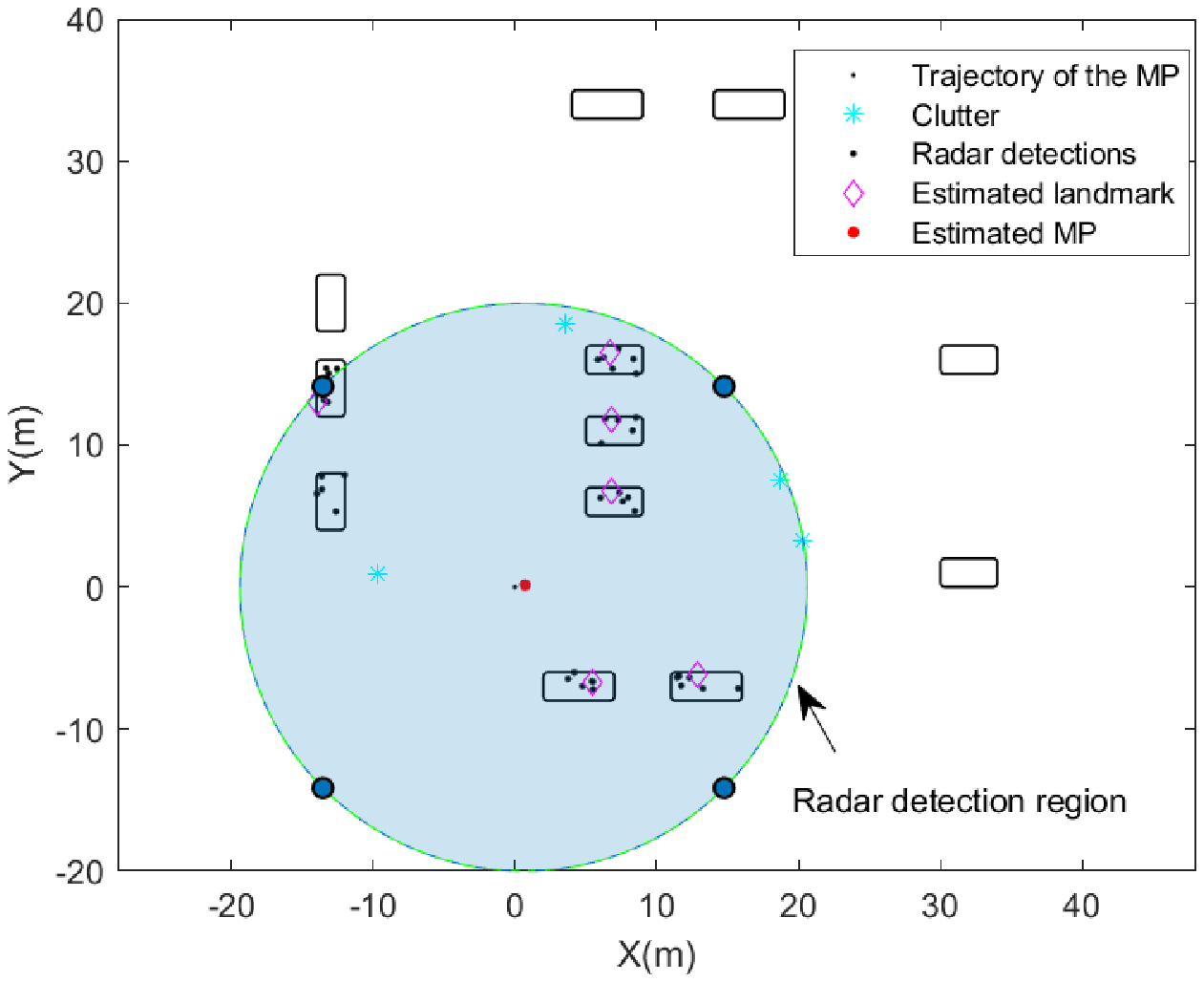}\label{subfig:exp1}} \hfil
    \subfloat[$k=14$]{\includegraphics[clip, trim=7mm 1mm 14mm 6mm,width=0.4\linewidth]{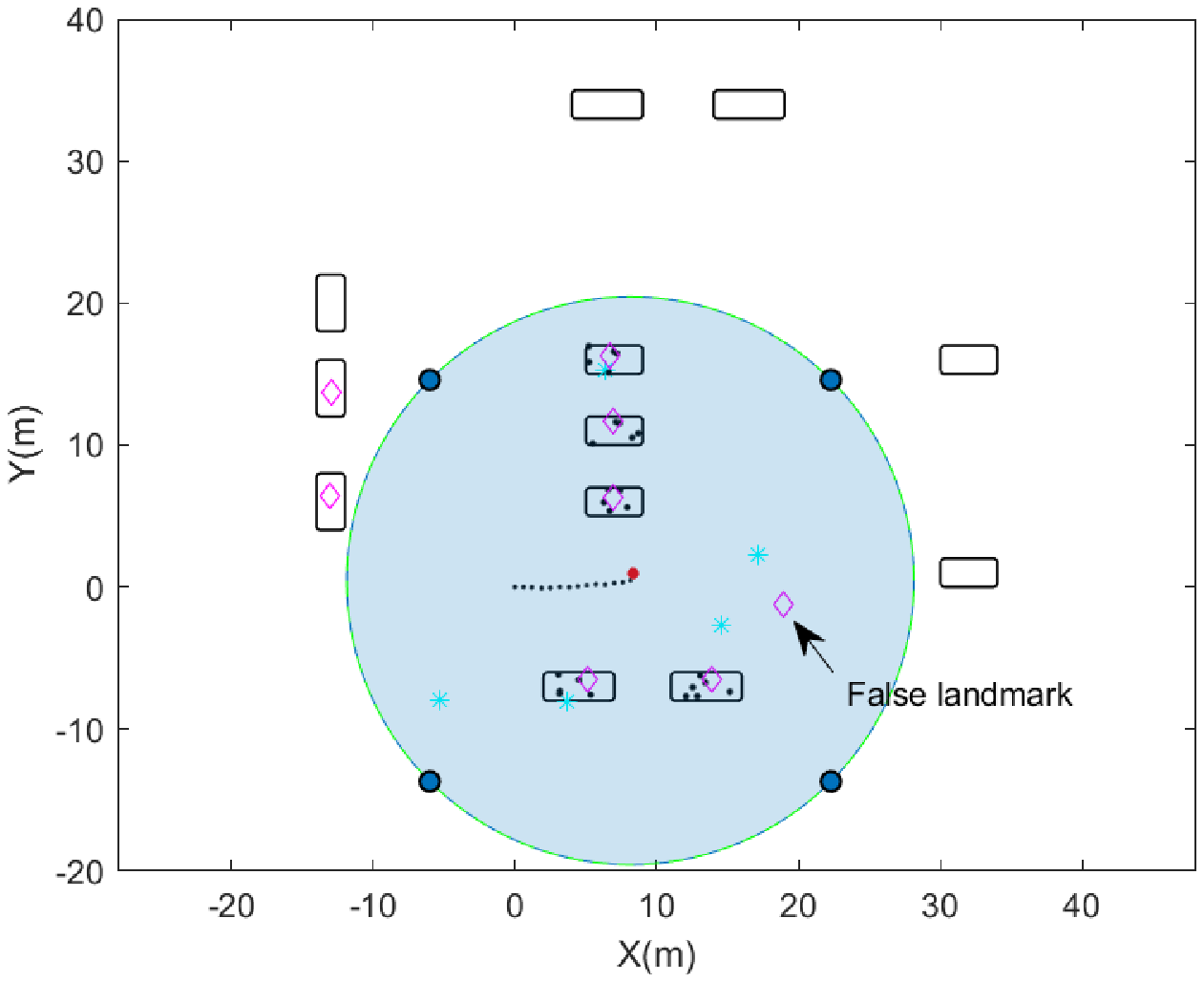}\label{subfig:exp2}} \vspace*{-1mm}
    
    \subfloat[$k=25$]{\includegraphics[clip, trim=7mm 1mm 14mm 6mm,width=0.4\linewidth]{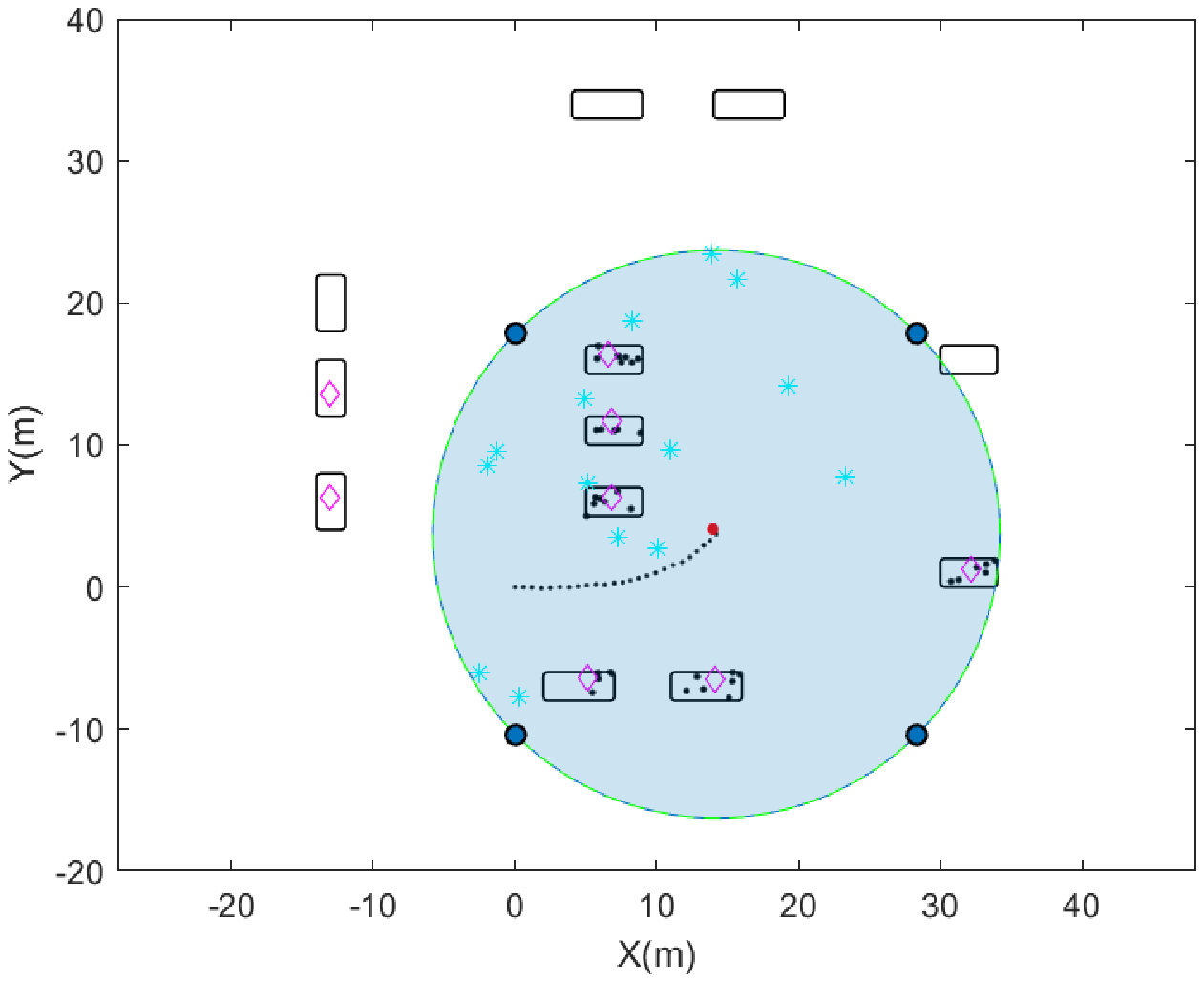}\label{subfig:exp3}} \hfil
    \subfloat[$k=40$]{\includegraphics[clip, trim=7mm 1mm 14mm 6mm,width=0.4\linewidth]{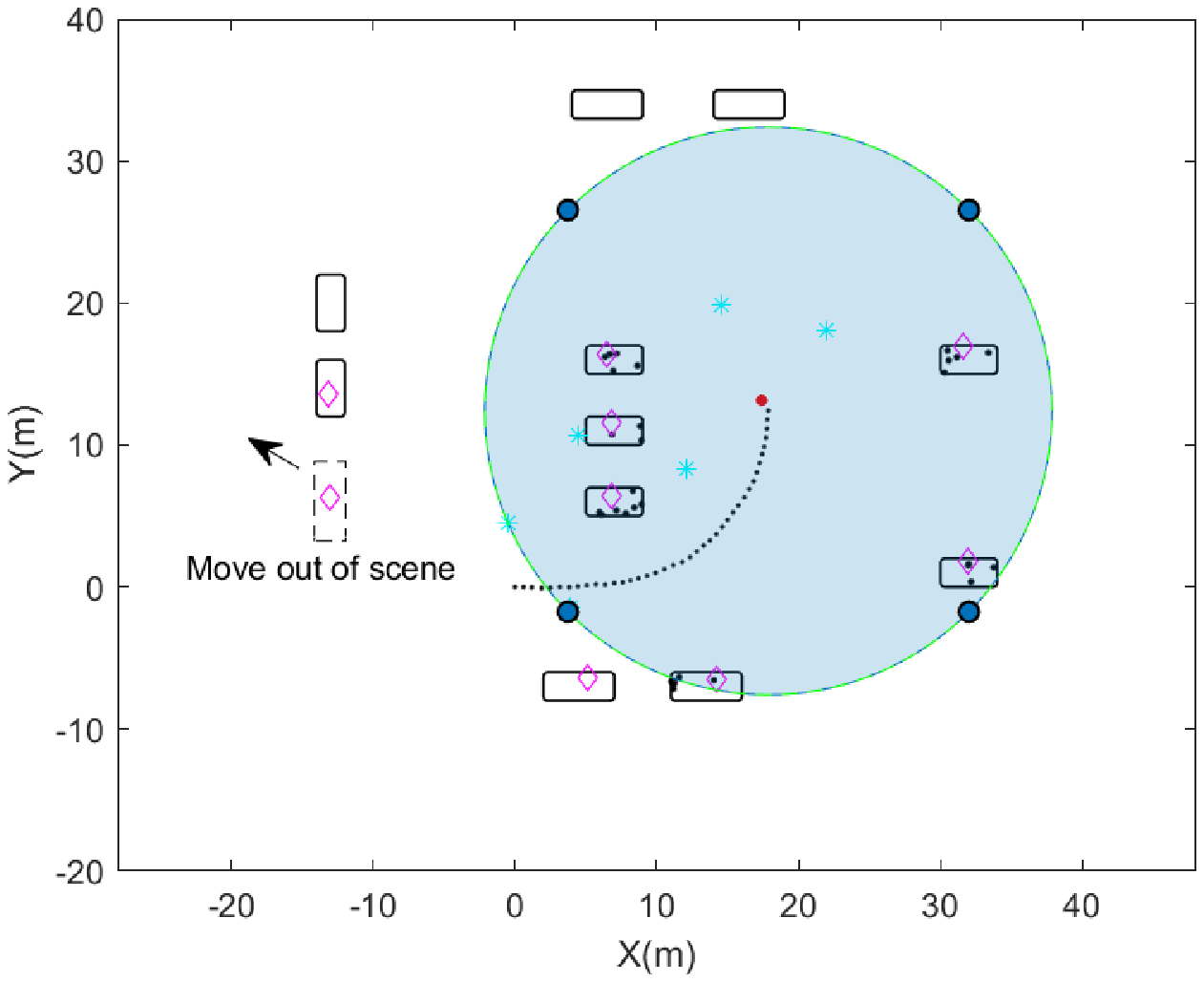}\label{subfig:exp4}}\vspace*{-1mm}
    
    \subfloat[$k=85$]{\includegraphics[clip, trim=7mm 1mm 14mm 6mm,width=0.4\linewidth]{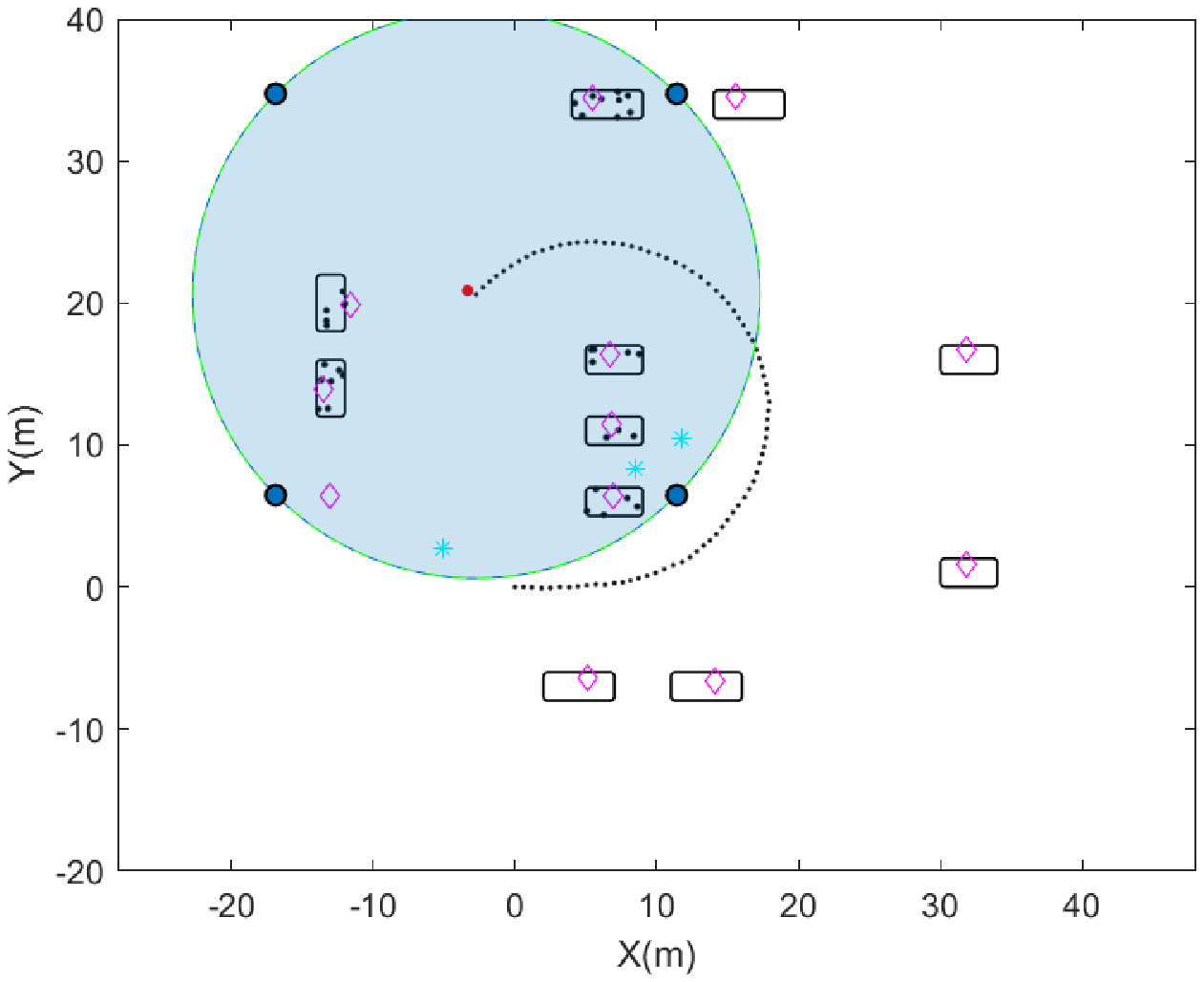}\label{subfig:exp5}} \hfil
    \subfloat[$k=110$]{\includegraphics[clip, trim=7mm 1mm 14mm 6mm,width=0.4\linewidth]{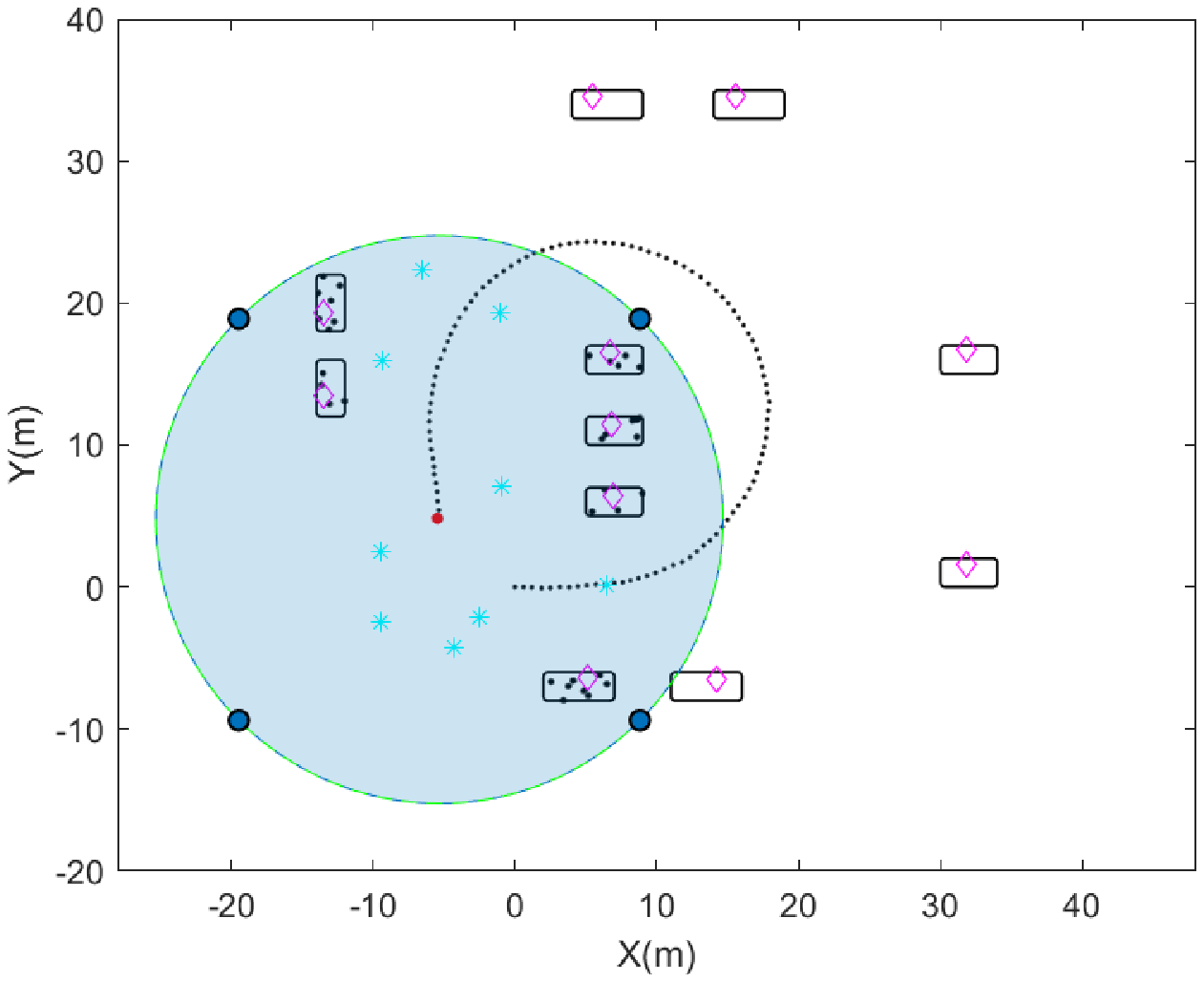}\label{subfig:exp6}}
    \caption{Illustration of the results of the proposed landmark management scheme using the radar EKF-SLAM algorithm. The simulated environment equates to a car park with the rectangles representing the parked cars. }
    \label{fig:experiment}
\end{figure*}

%===============================================================================
To assess the performance of our proposed landmark management scheme we run the EKF-SLAM algorithm in a simulated car park environment. The environment consists of several parked vehicles as shown in \autoref{subfig:exp1}. The cars are represented by rectangles and multiple radar detections are drawn uniformly from each rectangle. During the simulation the mobile platform moves along a designated trajectory, at a speed of roughly $4$~m/s for $k = 120$ time steps with a time interval of $\Delta t = 0.16$~seconds. The main parameters used in the algorithm are provided in \autoref{table:para}. We also provide sample Matlab code on github: \url{https://github.com/shuai000/SLAM_LandmarkManagement}.

\begin{table}[tb]
    \caption{Key parameters used in the simulation.}
    \ra{1.5}
    \centering 
    \begin{tabular}{lc} 
        \toprule
        \textbf{Parameter} & \textbf{Value}  \\
        \midrule
        Radar detection range, $R_{\text{max}}$ & 20m \\
        \midrule
        Measurement noise covariance, $R_k$ & $\begin{bmatrix} 0.5^2 & 0 \\ 0 & (1 \times \frac{\pi}{180})^2 \end{bmatrix}$ \\
        Process noise covariance, $Q_k$ & $\begin{bmatrix} 1.5e^{-3} & 0 & 0\\ 0 & 1.5e^{-3} & 0 \\ 0 & 0 & 5e^{-5}\end{bmatrix}$ \\
        Odometer noise covariance, $U_k$ & $\begin{bmatrix} 0.02^2 & 0 \\ 0 & (0.008 \times \frac{\pi}{180})^2 \end{bmatrix}$ \\ 
        \midrule
        Threshold $\alpha$ & 500 \\           
        Threshold $\beta$ & 20 \\        
        Clustering threshold $\gamma_c$ & 2.5m \\
        Sifting threshold $\gamma_s$ & 3m \\
        Cluster association threshold $\gamma_a$ & 3.5m \\
        Landmark merging threshold $\gamma_m$ & 1.5m \\
        Minimum detection points $N_{c_1}$ & 6 \\
        Minimum clustering points $N_{c_2}$ & 2 \\
        \midrule
        M/N logic (landmark initialization) & 5/3 \\
        M/N logic (landmark removal) & 10/2 \\ 
        \bottomrule
    \end{tabular}
    \label{table:para}
\end{table}

Examples of several time points from a single simulation are shown in \autoref{fig:experiment}. Due to a relatively high clutter rate, at $k=14$ a landmark is mistakenly confirmed. As this landmark is not consistently associated with radar detections, by $k=25$ this false landmark has been removed by the M/N landmark removal logic. New landmarks are confirmed and registered in the state vector when they are within radar detection range, and meet the landmark confirmation rule. For example, at $k=25$ we can see there is a new landmark which has been confirmed in the system. At $k=40$, one of the vehicles (registered in the state vector at $k=4$) drives out of scene. The landmark management scheme is not aware of this change until at $k=85$ the location of the ``disappeared'' vehicle is within the radar detection range. Since the expected detections from this registered landmark are not observed, this landmark is removed  according to the M/N landmark removal logic.

Next, we statistically evaluate the performance of the proposed solution using Monte Carlo simulations. The simulation is designed with two scenarios, one with a low clutter rate and the other with a high clutter rate. The results for $100$ Monte Carlo simulations are shown in \autoref{table:MC}. The table lists the average root mean square error (RMSE) for both platform pose and landmark position estimation. For the landmark management, we provide the mean delay for new landmark inclusion and landmark removal. We also include the mean number of false and missed landmarks per simulation along with the maximum number of false and missed landmarks observed in a single simulation. Based on these results, we conclude that: 1) the proposed rule based landmark scheme is able to efficiently conduct landmark initialization, maintenance, removal and merging; 2) the rule based landmark management scheme can be easily incorporated into the existing EKF-based SLAM framework. 

\begin{table}[tb]
    \caption{Monte Carlo simulation results} 
    \centering 
    \ra{1.2}
    \begin{tabular}{lcc} 
        \toprule 
        \textbf{Metric} & \textbf{Low Clutter} & \textbf{High Clutter} \\
        \midrule
        Platform position avg. RMSE (m) & 0.81 & 0.90 \\
        Platform heading avg. RMSE (deg) & 3.26  & 3.50 \\
        Landmark estimation MAE (m) & 1.23 & 1.34 \\
        \midrule
        Landmark inclusion mean delay & 2.45 & 3.22 \\ 
        Landmark removal mean delay & 10.85 & 11.00 \\
        \midrule
        Mean (Max) false landmarks & 0.13 (4)  & 3.02 (7) \\
        Mean (Max) missed landmarks & 0.2 (4) & 0.23 (5) \\
        \bottomrule
    \end{tabular}
    \label{table:MC} 
\end{table}

%===============================================================================
\section{Conclusions}
\label{sec:conclusion}
%===============================================================================
In this paper we proposed a rule based landmark management scheme which can be easily incorporated into existing SLAM frameworks and was demonstrated using the EKF-SLAM algorithm. A detailed discussion of the implementation of the state prediction and update, as well as landmark removal and inclusion are provided. Simulation results show that the proposed solution can effectively manage the landmarks while the mobile platform is conducting a SLAM task. This potentially offers promising results for practical vehicular based applications such as autonomous driving.  

%===============================================================================
\appendices
\section{EKF Prediction} 
\label{ap:ekf_predict}
%===============================================================================
For the prediction of the system state in the EKF, unlike the standard linear Kalman filter, the state vector (mean) is no longer independent of the state covariance. They are linked via the Jacobians $\nabla f_{\x^a_k}$ and $\nabla f_{\bu_k}$ which are obtained via the partial differentiation of $f$ w.r.t. $\x^a$ and $\bu$, taken at the point $(\x_{k-1|k-1}^a,\ \bu_k)$ giving
\begin{align*}
	\nabla f_{\x^a_k} &= \left.\dfrac{\partial f}{\partial \x^a_k}\right|_{(\x_{k-1|k-1}^a,\ \bu_k)} \\ 
	&=\left[\begin{array}{c c c |c }
	    1 & 0 & -\Delta t v_k A & \bm{0}_{1 \times 2N_k} \\ 
	    0 & 1 & \Delta t v_k B & \bm{0}_{1 \times 2N_k} \\ 
		0 & 0 & 1 & \bm{0}_{1 \times 2N_k} \\ 
		\midrule
		\bm{0}_{ 2N_k \times 1} & \bm{0}_{ 2N_k \times 1} & \bm{0}_{ 2N_k \times 1} & I_{2N_k \times 2N_k}
	\end{array}\right],
\end{align*}
\begin{align*}
	\nabla f_{\bu_k} &= \left.\dfrac{\partial f}{\partial \bu_k}\right|_{(\x_{k-1|k-1}^m,\ \bu_k)} \\ 
	&= \left[\begin{array}{c c c |c }
		\Delta t B & \frac{-\Delta t ^2 }{2} v_k A \\ 
		\Delta t A & \frac{\Delta t ^2 }{2} v_k B \\
		0 & \Delta t \\ 
		\midrule
		\bm{0}_{ 2N_k \times 1} & \bm{0}_{ 2N_k \times 1}
	\end{array}\right],
\end{align*}
where
\begin{align*}
    A &= \sin\left(\theta_{k-1|k-1} + \frac{\Delta t \psi_k}{2}\right),\\
    B &= \cos\left(\theta_{k-1|k-1} + \frac{\Delta t \psi_k}{2}\right).
\end{align*}
Note that there is no dimension change during state prediction. The size of $\nabla f_{\x^a}$ is $N_{k-1}^a \times N_{k-1}^a$ where $N_{k-1}^a$ is the size of the augmented state vector at time $k-1$ and the size of $\nabla f_{\bu_k}$ is $N_{k-1}^a \times 2$. $I_{2N_k}$ denotes the identity matrix with size $2N_k \times 2N_k$.

%===============================================================================
\section{Likelihood \& EKF Update} 
\label{ap:ekf_update}
%===============================================================================
The likelihood is defined as
\begin{align*}
    \Lambda (\z_k | \x_{k|k-1}^m,\, \p_{k-1}^n) = \frac{1}{\sqrt{(2\pi)^2 |S_k|}} \exp{\left(\hspace*{-0.7mm}-\frac{1}{2}\e_{n,k}^T S_k^{-1} \e_{n,k}\hspace*{-0.3mm}\right)},
\end{align*}
with $\e_{n,k} = \z_k - h_n(\x_{k|k-1}^m)$, where following the standard EKF update procedure, 
\begin{align*}
    S_k &= \nabla h_{\x_k^a}^n P_{k|k-1}^a \nabla {h_{\x_k^a}^n}^T + R_k,\\
	W_k & = P_{k|k-1}^a \nabla {h_{\x_k^a}^n}^T S_k^{-1},
\end{align*}
and the partial differentiation matrix is given by
\begin{align*}
    \nabla h_{\x_k^a}^n &= \left.\dfrac{\partial h_n}{\partial \x_k^a}\right|_{\x_{k|k-1}^a} \\ 
	&= \begin{bmatrix}
	    -\frac{\Delta x}{r_n} & -\frac{\Delta y}{r_n}  & 0 & \cdots & \frac{\Delta x}{r_n} & \frac{\Delta y}{r_n} & \cdots \\[1mm] 
		\frac{\Delta y}{r_n^2} & -\frac{\Delta x}{r_n^2} & -1 & \cdots & -\frac{\Delta y}{r_n^2} & \frac{\Delta x}{r_n^2} & \cdots  
	\end{bmatrix},
\end{align*}
with
\begin{align*}
    \Delta x & = p_k^n(x) - x_{k|k-1} \\
	\Delta y & = p_k^n(y) - y_{k|k-1} \\
	r_n & = \sqrt{(\Delta x)^2 + (\Delta y)^2}.
\end{align*}

%===============================================================================
\section{Landmark Inclusion} 
\label{ap:inclusion}
%===============================================================================
The augmented state can be expressed as a function of the original state and the cluster centers,
\begin{align*}
    \x_{k|k}^{a,j} &= g(\x_{k|k}^{a,j-1}, \bc_k^j) \\
    &=\left[\begin{array}{c}
	    \x_{k|k}^{a,j-1} \\ 
	    \midrule
		x_{k|k} + r_k^j \cos(\theta_{k|k} + \phi_k^j) \\ 
		y_{k|k} + r_k^j \sin(\theta_{k|k} + \phi_k^j)
	\end{array}\right],
\end{align*}
with the Jacobin matrix defined as,
\begin{align*}
    J_1 &= \left.\dfrac{\partial g}{\partial \x^{a, j-1}}\right|_{\x=\x_{k|k}^{a,j-1}, \bc=\bc_k^j} \\ 
    & = \left[\begin{array}{cccc}
	    &   & I_{N \times N} &  \\ 
	    \midrule  
	    1 & 0 & -r_k^j \sin (\theta_k + \phi_k^j) & \bm{0}_{1 \times (N-3)} \\
	    0 & 1 &  r_k^j \cos (\theta_k + \phi_k^j)  & \bm{0}_{1 \times (N-3)}
	\end{array}\right], \\
	J_2 &= \left.\dfrac{\partial g}{\partial \bc_k^j}\right|_{\x=\x_{k|k}^{a,j-1}, \bc=\bc_k^j} \\ 
	& = \left[\begin{array}{cc}
	    \bm{0}_{N \times 1} & \bm{0}_{N \times 1} \\
	    \midrule 
	    \cos (\theta_k + \phi_k^j) & -r_k^j \sin (\theta_k + \phi_k^j) \\
		\sin (\theta_k + \phi_k^j) & r_k^j \cos (\theta_k + \phi_k^j)  
		\end{array}\right].
\end{align*}

\printbibliography

@article{Granstrom2016ExtendedTargetTracking_Arxiv,
  title={Extended object tracking: Introduction, overview and applications},
  author={Granstrom, Karl and Baum, Marcus and Reuter, Stephan},
  journal={Journal of Advances in Information Fusion},
  volume={12},
  number={2}, 
  pages={139--174},
  year={2017},
  eprint={1604.00970},
  eprinttype={arxiv},
}

@article{lee2020EKF_SLAM_TAES,
  title={Experimental results and posterior cram{\'e}r--rao bound analysis of {EKF}-based radar {SLAM} with odometer bias compensation},
  author={Lee, Hyukjung and Chun, Joohwan and Jeon, Kyeongjin},
  journal={IEEE Transactions on Aerospace and Electronic Systems},
  volume={57},
  number={1},
  pages={310--324},
  year={2020},
  doi={10.1109/TAES.2020.3016873}
}

@article{Dissanayake2002MapManagement_Robots,
  title={Map management for efficient simultaneous localization and mapping ({SLAM})},
  author={Dissanayake, Gamini and Williams, Stefan B and Durrant-Whyte, Hugh and Bailey, Tim},
  journal={Autonomous Robots},
  volume={12},
  number={3},
  pages={267--286},
  year={2002},
  doi={10.1023/A:1015217631658}
}

@article{Grisetti2007Grid_mapping_SLAM_TransRobotics,
  title={Improved techniques for grid mapping with {Rao-Blackwellized} particle filters},
  author={Grisetti, Giorgio and Stachniss, Cyrill and Burgard, Wolfram},
  journal={IEEE Transactions on Robotics},
  volume={23},
  number={1},
  pages={34--46},
  year={2007},
  doi={10.1109/TRO.2006.889486}
}

@phdthesis{Bailey2002SLAM_fundamentals,
  title={Mobile robot localisation and mapping in extensive outdoor environments},
  author={Bailey, Tim},
  year={2002},
  institution={University of Sydney}
}

@ARTICLE{SLAM_past_future_2016,
  author={Cadena, Cesar and Carlone, Luca and Carrillo, Henry and Latif, Yasir and Scaramuzza, Davide and Neira, Jos\'e and Reid, Ian and Leonard, John J.},
  journal={IEEE Transactions on Robotics}, 
  title={Past, Present, and Future of Simultaneous Localization and Mapping: Toward the Robust-Perception Age}, 
  year={2016},
  volume={32},
  number={6},
  pages={1309--1332},
  doi={10.1109/TRO.2016.2624754}
}

@ARTICLE{SLAM_survey_2016,
  author={Bresson, Guillaume and Alsayed, Zayed and Yu, Li and Glaser, Sébastien},
  journal={IEEE Transactions on Intelligent Vehicles}, 
  title={Simultaneous Localization and Mapping: A Survey of Current Trends in Autonomous Driving}, 
  year={2017},
  volume={2},
  number={3},
  pages={194--220},
  doi={10.1109/TIV.2017.2749181}
}

@article{VivetFMCW_SLAM_2013,
  title={Localization and mapping using only a rotating {FMCW} radar sensor},
  author={Vivet, Damien and Checchin, Paul and Chapuis, Roland},
  journal={Sensors},
  volume={13},
  number={4},
  pages={4527--4552},
  year={2013},
  doi={10.3390/s130404527}
}

@article{Lee2013PhD_slam,
  title={SLAM with dynamic targets via single-cluster {PHD} filtering},
  author={Lee, Chee Sing and Clark, Daniel E and Salvi, Joaquim},
  journal={IEEE Journal of Selected Topics in Signal Processing},
  volume={7},
  number={3},
  pages={543--552},
  year={2013},
  doi={10.1109/JSTSP.2013.2251606}
}

@article{2021Landmark_elimination,
  title={Sensor Fusion-Based Approach to Eliminating Moving Objects for {SLAM} in Dynamic Environments},
  author={Dang, X. and Rong, Z. and Liang, X.},
  journal={Sensors},
  volume={21},
  number={1:230},
  year={2021},
  doi={10.3390/s21010230}
}

@article{Vu2011gridSLAM,
  title={Grid-based localization and local mapping with moving object detection and tracking},
  author={Vu, Trung-Dung and Burlet, Julien and Aycard, Olivier},
  journal={Information Fusion},
  volume={12},
  number={1},
  pages={58--69},
  year={2011},
  doi={10.1016/j.inffus.2010.01.004}
}

@INPROCEEDINGS{Car_tracking_2014,
  author={Granstr\"om, Karl and Reuter, Stephan and Meissner, Daniel and Scheel, Alexander},
  booktitle={17th International Conference on Information Fusion (FUSION)}, 
  title={A multiple model {PHD} approach to tracking of cars under an assumed rectangular shape}, 
  year={2014},
  pages={1--8}
}

@article{Bailey2006SLAM,
  title={Simultaneous localization and mapping ({SLAM}): Part {II}},
  author={Bailey, Tim and Durrant-Whyte, Hugh},
  journal={IEEE Robotics \& Automation Magazine},
  volume={13},
  number={3},
  pages={108--117},
  year={2006},
  doi={10.1109/MRA.2006.1678144}
}

@inproceedings{Wang2003SLAMwithDTMO,
  title={Online simultaneous localization and mapping with detection and tracking of moving objects: Theory and results from a ground vehicle in crowded urban areas},
  author={Wang, Chieh-Chih and Thorpe, Charles and Thrun, Sebastian},
  booktitle={IEEE International Conference on Robotics and Automation},
  volume={1},
  pages={842--849},
  year={2003},
  doi={10.1109/ROBOT.2003.1241698}
}

@article{Fazekas2021OdometryModelVehicle,
  title={Calibration and improvement of an odometry model with dynamic wheel and lateral dynamics integration},
  author={Fazekas, M{\'a}t{\'e} and G{\'a}sp{\'a}r, P{\'e}ter and N{\'e}meth, Bal{\'a}zs},
  journal={Sensors},
  volume={21},
  number={2:337},
  year={2021},
  doi={10.3390/s21020337}
}

@article{Li2017FusionSLAMVehicles,
  title={Hybrid filtering framework based robust localization for industrial vehicles},
  author={Li, Liang and Yang, Ming and Wang, Chunxiang and Wang, Bing},
  journal={IEEE Transactions on Industrial Informatics},
  volume={14},
  number={3},
  pages={941--950},
  year={2017},
  doi={10.1109/TII.2017.2738016}
}

@article{Guivant2001optimization_EKF_SLAM,
  title={Optimization of the simultaneous localization and map-building algorithm for real-time implementation},
  author={Guivant, Jose E and Nebot, Eduardo Mario},
  journal={IEEE Transactions on Robotics and Automation},
  volume={17},
  number={3},
  pages={242--257},
  year={2001},
  doi={10.1109/70.938382}
}

@online{hong2021radar_SLAM,
  title={Radar {SLAM}: A robust slam system for all weather conditions},
  author={Hong, Ziyang and Petillot, Yvan and Wallace, Andrew and Wang, Sen},
  year={2021},
  eprint={2104.05347},
  eprinttype={arxiv},
}

@INPROCEEDINGS{Ester96adensity-based,
  author = {Martin Ester and Hans-Peter Kriegel and Jörg Sander and Xiaowei Xu},
  title = {A density-based algorithm for discovering clusters in large spatial databases with noise},
  booktitle = {Proceedings of the Second International Conference on Knowledge Discovery and Data Mining},
  year = {1996},
  pages = {226--231},
  doi={10.1.1.121.9220}
}

\end{document}